\documentclass{article}
\usepackage{spconf}
\usepackage{graphicx}
\usepackage{amsmath}
\usepackage{amssymb}
\usepackage{times}
\usepackage{epsfig}
\usepackage{wrapfig}
\usepackage{array}
\usepackage[final]{pdfpages}
\usepackage{multirow}
\usepackage{multicol}
\usepackage{caption}
\usepackage{subcaption}
\usepackage{array}
\usepackage{tabularx}
\graphicspath{{./images/}}
\usepackage[pagebackref=true,breaklinks=true,letterpaper=true,colorlinks,bookmarks=false]{hyperref}

\title{Fine-to-Coarse Knowledge Transfer for Low-Res Image Classification}

\name{Xingchao Peng$^{\star}$ \qquad Judy Hoffman $^{\dagger}$ \qquad Stella X. Yu$^{\ddagger}$ \qquad Kate Saenko$^{\star}$}

\address{$^{\star}$ {\small{Computer Science Department, Umass Lowell}} \\
    $^{\dagger}$ {\small{Electronic Engineering and Computer Science Department, UC Berkeley}} \\
    $^{\ddagger}$ \small{International Computer Science Institute, UC Berkeley}
}

\begin{document}

\maketitle

\begin{abstract}

We address the difficult problem of distinguishing fine-grained object categories in low resolution images. We propose a simple an effective deep learning approach that transfers fine-grained knowledge gained from high resolution training data to the coarse low-resolution test scenario.
Such fine-to-coarse knowledge transfer has many real world applications, such as identifying objects in surveillance photos or satellite images where the image resolution at the test time is very low but 
plenty of high resolution photos of similar objects are available. Our extensive experiments on
two standard benchmark datasets containing fine-grained car models and bird species demonstrate that our approach can effectively transfer fine-detail knowledge to  coarse-detail imagery.
\end{abstract}

\begin{keywords}
Fine-grained Classification, Low Resolution, Deep Learning
\end{keywords}

\section{Introduction}
\label{introduction}

\begin{figure}[t]
    \centering
    \includegraphics[width=0.9\linewidth ]{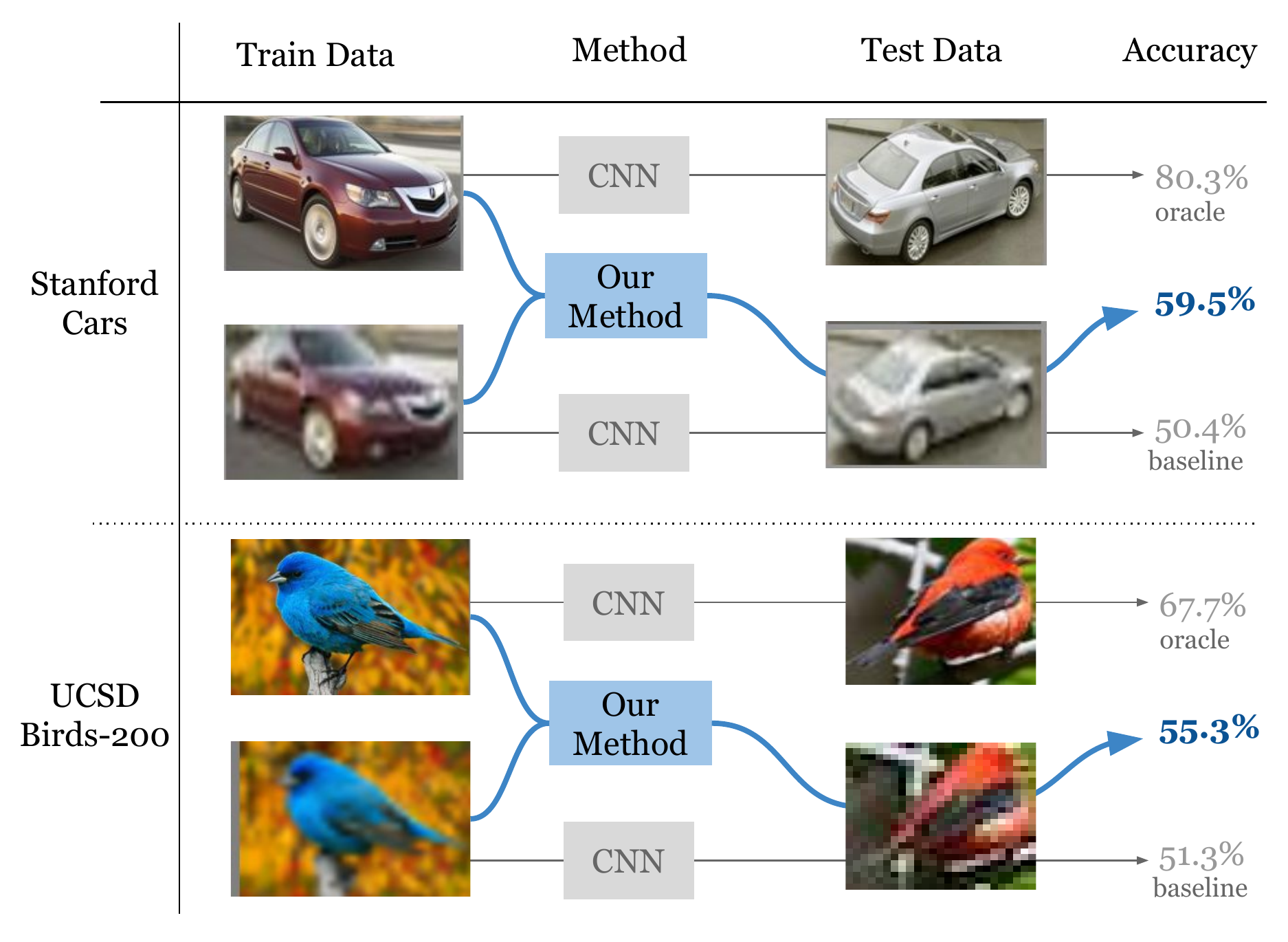}
    \vspace{-3mm}
    \caption{Fine-grained category classification, such as classifying a car's make and model, or a bird's species, is extremely challenging in coarse, low-resolution images. We propose a ``Staged Training'' approach for deep convolutional neural networks that significantly improves classification by transferring knowledge from high-resolution training data.}
    \label{fig:task}
    \vspace{-0.2in}
\end{figure}

Fine-grained classification methods must distinguish between very similar categories, such as the make and model of a car (Toyota Corolla vs Nissan Leaf) or the species of a bird (Indigo Bunting vs Blue Grosbeak). This requires learning subtle discriminative features, for example, the car manufacturer logo, or the special patterns on a bird's beak. However, such features are challenging to extract when test images are coarse and have low effective resolution (see Figure~\ref{fig:task}). We ask, is it still possible to rely on fine details to identify the category of interest as these details become blurred and diminished?

Existing approaches to fine-grained classification~\cite{lin2015bilinear,zhang2014part} use convolutional neural networks (CNNs) to learn
such discriminative feature representations. Visualizations~\cite{viscnn,RCNN} have shown that middle layers of CNNs give rise to features such as logos or object parts, while higher layers capture overall object configuration. However these methods typically assume that both the training and test images are sufficiently high-res (e.g., 227-by-227 pixels). In real world applications, images of test objects can be much smaller, e.g., 50-by-50 pixels or less, or could have low effective resolution due to blurring, lighting or other effects. Models trained on high-res data fail miserably in these scenarios due to the considerable appearance shift between training and test data. On the other hand, training on matched low-res data results in representations that lack discriminability and obtain inferior accuracy.

In this paper, we show that it is possible to transfer the knowledge about discriminative features from the high-res domain to the low-res domain and significantly improve accuracy. Our assumption is that high-resolution labeled data is available for training, while at test time only low-resolution data is given.  We propose a simple staged training procedure that first trains the representation on high-res data, learning discriminative mid-level features. It then artificially lowers the resolution of the training data to match the test domain, and continues fine-tuning the representation, adapting discovered discriminative features to the target resolution.

We compare our approach to conventional training, and also to traditional methods for super-resolution. Super-resolution attempts to improve the quality of low resolution images, for example, through the application of cross-scale patch redundancy~\cite{sr}. 
One approach is to use super-resolution on the low-res images before applying the classifier network. 
As we show, super-resolution approaches complex and time consuming compared to our method, and cannot handle the large variations in resolution that occur in practical scenarios.
Through extensive experiments on the Stanford Cars~\cite{stanfordcar} and Caltech-UCSD Birds (UCB-200-2011)~\cite{wah2011caltech} datasets, we demonstrate the advantage of our approach over existing methods.

\section{Related Work}
\label{related}

The traditional approach to recovering high frequency details from a low-resolution image is referred to as the ``super resolution'' method. Example-based super-resolution methods \cite{srregistration, srfast} work on a set of low-resolution images of the same scene. More recent work leveraged  sophisticated  methods to recover the lost details from a single image. \cite{sr} proposed a unified framework to employ in-scale patch redundancy and cross-scale patch redundancy, based on the observation that patches in natural images tend to redundantly recur, both at the same scale and at different scales. \cite{li2010low} proposed a learning-based approach to improve low-resolution face recognition performance with locality preserving mappings. These approaches are heavily dependent on repeated information in a single image or across a batch of images, and do not focus on classification. In contrast, our method directly improves fine-grained classification performance by utilizing rich discriminative information in fine-scale training images.

Fine-grained classification distinguishes subcategories of objects within a single basic-level category,
and has been the focus of much research. Applications have included natural objects like animal or plant species~\cite{farrell2011birdlets, khosla2011novel, angelova2013image, belhumeur2008searching}, or man-made objects~\cite{stanfordcar, maji2013fine}. \cite{farrell2011birdlets, zhang2014part}  find parts of the object and align object pose, while \cite{stanfordcar} proposes to utilize the 3D shape of cars to perform fine-grained car classification. These works are all based on an ideal assumption that all images are high quality. In contrast, \cite{chevalierlr} performs fine-grained classification of man-made objects with different resolutions. 

However, their model is a CNN designed and trained exclusively on low-res images, while our approach effectively transfers knowledge from high-res training data to the low-res domain. 

\section{Fine-to-Coarse Knowledge Transfer via Staged Training}
\label{method}

We propose a simple but effective knowledge transfer approach that improves fine-grained category classification in very low resolution images.
We assume that, even though the test data has low resolution, we have access to high resolution labeled training data. This is a reasonable assumption as it is much easier to label subcategories in high-res data, and most existing datasets are high-res. We aim to transfer knowledge from such datasets to real world scenarios that lack resolution.

The basic intuition behind our approach is to utilize high-quality distinguishing details in the training domain to guide representation learning for the target low-res domain. Conventional wisdom dictates that machine learning models should be trained on data that is as similar to the test data as possible, otherwise the mismatch in input features leads to a drop in performance~\cite{saenko2010adapting}. Our experiments support this by showing that CNNs trained in the traditional way on high-res data fail miserably on low-res inputs. However, training on matched low-res data also leads to low performance as discriminative features are lost. 

\begin{figure}
    \centering
    \includegraphics[width=0.9\linewidth]{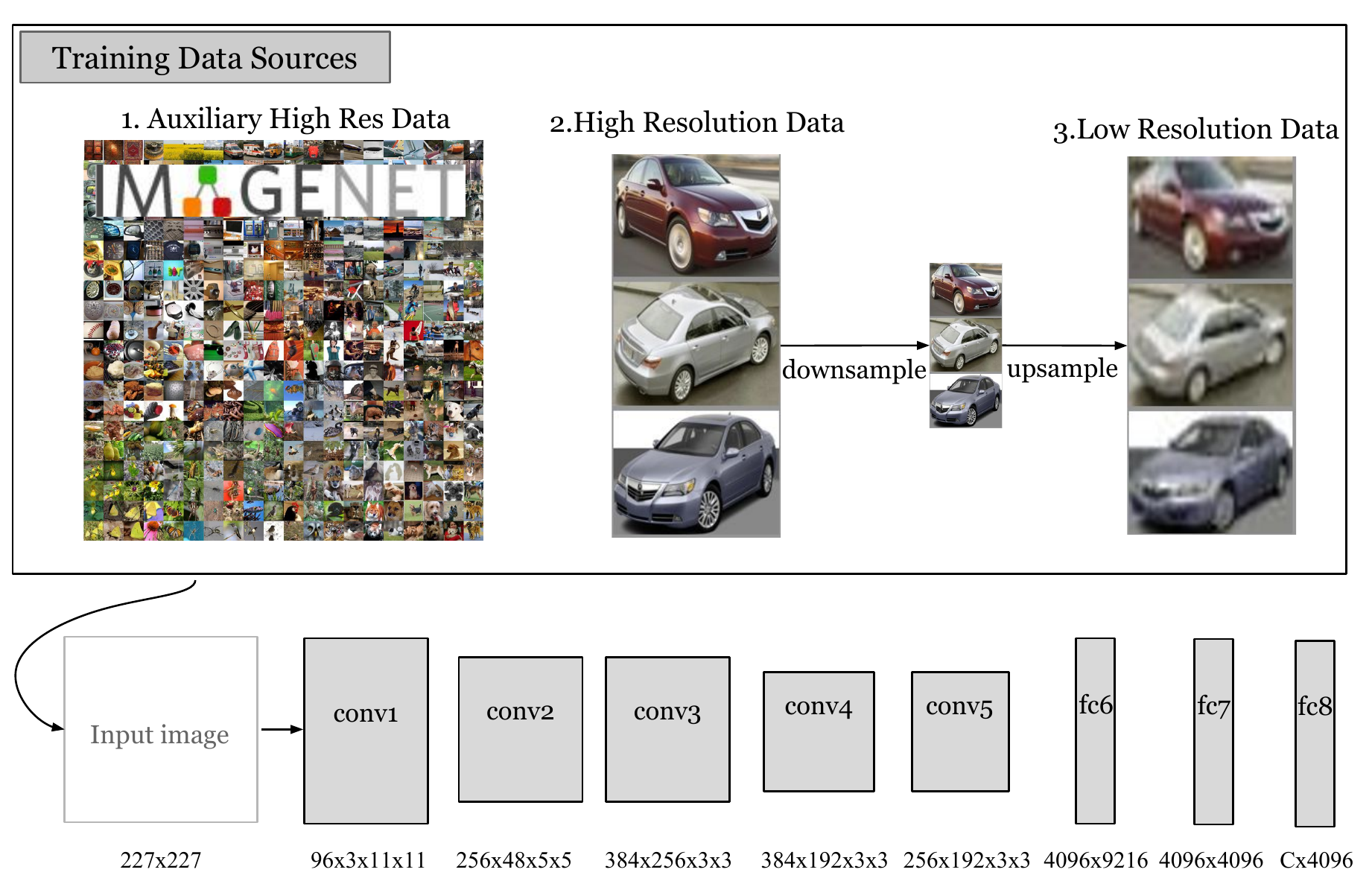}
    \vspace{-2mm}
    \caption{Our staged training procedure using the ``AlexNet'' architecture~\cite{alexnet}. 1) Pre-train using a large high resolution auxiliary data source (ex: ImageNet~\cite{ilsvrc}). 2) Fine-tune on our domain-specific high res data. 3) Fine-tune on artificial low resolution (downsampled and then upsampled) in-domain data.
    }
    \label{fig:overview}
    \vspace{-0.2in}
\end{figure}

Inspired by domain adaptation and transfer learning approaches~\cite{saenko2010adapting,sun2016return,peng2015learning}, we design an adaptive training procedure that consists of the following stages: First, we initialize the model by training on a large auxiliary dataset, then continue to fine-tune it (train with a lower learning rate) on the high-res fine-grained category training data. We then artificially reduce the effective resolution of the training data to match that of the target domain and continue fine-tuning on this data, adapting the representation to the low-res domain. Our visualizations  of the resulting features (Sec.~\ref{sec:cars}) show that this staged training procedure results in stronger discriminative feature activations on the target low-res domain. An overview of the approach is shown in Fig~\ref{fig:overview}.

\vspace{0.05in}
\noindent \textbf{Convnet Architecture} In this paper we use the architecture proposed by~\cite{alexnet}, commonly known as ``Alexnet''.  It has five convolutional layers, three fully connected layers, including a 1000-dimensional output layer, and has over 60 million parameters. To reduce overfitting, the ``Alexnet'' adopts ``dropout'' regularization method and to make training faster, it uses non-saturating neurons and a very efficient GPU implementation of the convolution operation. The input image size is 227-by-227 pixels. We pre-train the network on 1.2M labeled high-res examples in ImageNet~\cite{ilsvrc} (both basic category and subcategory) by downsampling them to the input size.

\vspace{0.05in}
\noindent \textbf{High-Res Training Stage} We initialize the network with the representation learned on ImageNet, transferring all layers except the output layer, i.e. \emph{conv1}$\scriptsize{\sim}$\emph{fc7}. We change the output \emph{fc8} layer from 1000 dimensions to the number of categories in the dataset and initialize the weights with a standard Gaussian distribution. We then continue training on the high-res fine-grained category data.

\vspace{0.05in}
\noindent \textbf{Low-Res Training Stage} At this stage, the high-res training data is first downsampled to the target domain resolution (50-by-50 in our experiments) and then up-sampled again to 227-by-227 to match the input size of the CNN. We assume that the target resolution is known. We then continue to fine-tune the representation on this data with a low learning rate. 

\vspace{0.05in}
\noindent \textbf{Visualizing Discriminative Features}
To analyse the effect of our staged training scheme, we devise a method for visualizing the resulting representation. Inspired by the feature heat-map method in \cite{viscnn}, we propose to visualize the most discriminative features learned for each category by our method and by the traditional method. We denote the whole pipeline of the convolutional neural network as a function $\nu=\Psi(I)$, where $\emph{I}$ is the input image and $\nu$ is the output vector from \emph{fc8}. For each pixel in the image $\emph{I}$, we gray out (set the value to be 128) a square patch centered at that pixel and render a new image $\emph{I}$'. We assign $\|\Psi(\emph{I})-\Psi(\emph{I'})\|^{2}$ to be the value of that pixel. After all the pixels in $\emph{I}$ get a value, we normalize to produce a heat map image. High heat map values then correspond to the most discriminative features, i.e. those that have the most effect on the predicted category.

\vspace{-0.1in}
\section{Experiments}
\label{experiments}
\vspace{-0.05in}

We evaluate on two fine-grained classification datasets:

\noindent {\bf Stanford Cars Dataset}~\cite{stanfordcar} was collected for fine-grained car classification. It contains 16,185 images of 196 classes of cars, which are at the level of Make, Model, Year. Most of the images are car-centered images and the average size of bounding boxes is 575-by-310. We follow the standard split of the dataset with 8144 training  and 8041 testing images. 

\noindent {\bf Caltech-UCSD Birds-200-2011 Dataset}~\cite{wah2011caltech} is a widely-used fine-grained classification benchmark with 11,788 images of 200 types of birds. These bird images are natural images taken in the wild, with bounding box size 260-by-235 on average, and are more likely to be coarse than the car images. We follow the standard train/test split.

We crop all training and testing images with the help of the known bounding box of the object,
and generate low-res data by downsampling to 50-by-50 pixels.

\vspace{-0.1in}
\subsection{Baselines}
\noindent \textbf{``AlexNet''} This baseline follows the traditional procedure and trains the network on the same resolution as the test data. For completeness, we show the results of testing on both low-res and high-res conditions.

\noindent \textbf{Mixed Training}   We also explore learning the filters from high-res images and low-res images at the same time. The high-res/low-res training images are combined the network is trained on the mixed data. During the test phase, we evaluate on high-res images and low-res images separately. 
Note that this mixed training scheme does not give preference to either resolution, unlike our adaptive method, which learns features that benefit the test domain.

\noindent \textbf{Super-Resolution} 
To compare with the traditional super-resolution method on fine-grained tasks, we apply the state-of-the-art Naive Bayes Super-Resolution Forest (NBSRF) proposed in \cite{Salvador2015} to up-scale all the low resolution training and testing images. The network is trained on the up-scaled images and tested on the up-scaled test set. 
We found that this works better than training on high-res and testing on up-scaled low-res, which leads to poor performance due to data mismatch.

 \vspace{-0.1in}
\subsection{Results on Stanford Car Dataset}\label{sec:cars}

\begin{table}[t]
\begin{center}
\begin{small}

\begin{tabular}{c|c|c|c}

\multirow{2}{*}{Train/Test Strategy} & \multirow{2}{*}{Train} &Test on&Test on\\ 
&&High-res&Low-res\\

\hline
``AlexNet''& High& 80.3 &1.7\\
``AlexNet'' & Low & 13.3 & 50.4\\
Mixed-Training & High+Low & 72.9 &59.3\\
BB-3D-G\cite{stanfordcar} &High & 67.6  &- \\
Super-Res NBSRF\cite{Salvador2015} & Low &-&50.8 \\
Staged-Training(L-H) & 1. Low 2. High & 65.2 & 18.1 \\
Staged-Training(H-L) & 1. High 2. Low &37.2 & \textbf{59.5}

\end{tabular}

\end{small}
\end{center}
\vspace{-4mm}
\caption{\textbf{Results on Stanford Cars Dataset} 
Accuracy for traditional training (``AlexNet''), several baseline methods, and our Staged-Training method. While we target the low-res test scenario, we also show results for high-res test for comparison. 
}
\label{tab:singleRes}
\vspace{-0.1in}
\end{table}

Table~\ref{tab:singleRes} summarizes the results. We see that learning filters directly on high-res images and testing on low-res images leads to 1.7\% accuracy, a huge drop from 50.44\% obtained by training on low-res images, which demonstrates the sensitivity of the CNN to the resolution domain mismatch. The  super resolution method NBSRF~\cite{Salvador2015} results in almost no boost. We also compared with the BB-3D-G~\cite{stanfordcar} method, which does not use CNNs and performs poorly compared to our CNN-based methods. ``Oracle'' Alexnet performance of training and testing on high-res reaches 80\% accuracy. Not surprisingly, training on low-res images and testing on high-res images leads to a low 13.31\%. 

We also implemented the LR-CNN structure proposed in \cite{chevalierlr}, except instead of contrast normalization we used local response normalization after the first convolutional layer. The accuracy obtained on the low-res Stanford Car Dataset was 19.1\%.

\begin{figure}[t]
\centering
\includegraphics[width=0.9\linewidth]{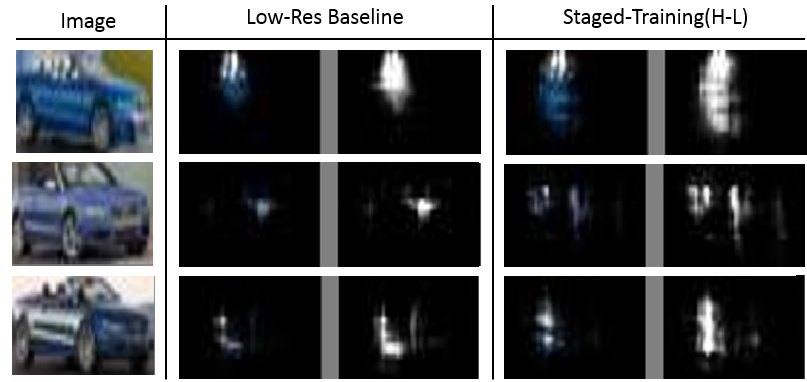}

\vspace{-3mm}
\caption{ 
We compare heat maps generated for traditional ``AlexNet'' (Low-Res Baseline) trained on low-res images with heat maps generated by our method (Staged-Training) for several low-res test images. Discriminative features learned by our method cover larger areas in the images and are stronger than those learned by the traditional baseline.
The heat-maps are generated with the method in Sec.~\ref{method}; we filter the original image with the heatmaps to render the discriminative patches.}
\label{fig:vis}
\vspace{-1em}
\end{figure}

Our proposed adaptive method, Staged-Training(H-L), improves low-res test accuracy from 50.44\% to 59.5\%, a surprising 18\% relative improvement.
For completeness, we apply our method in reverse and, as expected, transferring knowledge from low-res to high-res via Staged-Training hurts performance. We conclude that our approach is very effective at fine-to-coarse transfer.

To analyze why our approach renders such a high accuracy improvement, we use the method proposed in Section~\ref{method} to visualize the most discriminative feature heat map. In Figure~\ref{fig:vis}, we compare heat maps generated by ``AlexNet'' trained on low-res images with heat maps generated by our method for several low-res test images. From these (and other similar) visualizations we see that discriminative features learned by our method cover larger areas in the images and are stronger than those learned by the traditional baseline.

On this dataset, mixed training also considerably boosts performance to 59.33\%, on par with our method. We hypothesise that with mixed training, the network learns both fine-detail features and coarse features at the same time. However, this requires more parameters and thus more training data.  
To further investigate this, we  re-run experiments on varying amounts of training data, with results shown in Table \ref{tab:seqmix}. Here The results with $\dag$ and the results with * are comparable because they use exactly the same training and testing data. The  results show that staged training is better than mixed training when training data is limited. 

\begin{table}[t]

\begin{center}
\begin{small}
\begin{tabular}
{c|p{0.5cm} p{0.5cm} p{0.5cm} p{0.5cm} p{0.5cm} p{0.5cm} p{0.5cm} p{0.5cm}}
\hline
-&$L_{1k}$&$L_{2k}$&$L_{3k}$&$L_{4k}$&$L_{5k}$&$L_{6k}$&$L_{7k}$&$L_{8k}$ \\
\hline
$H_{0k}$&12.1&	19.4&	27.1&	29.8&	37.1&	41.6&	46.1&	50.4\\
$H_{1k}$&13.9 &21.6 &28.0 &33.7 &39.3 &45.2 &47.6 &51.6 \\
$H_{2k}$&17.2 &\textbf{24.3}\dag &32.6 &37.2 &42.8 &46.7 &50.1 &53.6 \\
$H_{3k}$&21.3 &28.9 &\textbf{34.6}\dag &40.5 &45.2 &50.6 &52.5 &55.6 \\
$H_{4k}$&22 &32.1 &37.2 &\textbf{42.4}\dag &46.5 &49.9 &53.6 &56.2 \\
$H_{5k}$&25.1 &33.4 &38.4 &44.9 &\textbf{47.6}\dag &51.9 &54.4 &57.1\\
$H_{6k}$&26.2 &33.8 &41.9 &46.6&50.0 &\textbf{52.8}\dag &54.9 &58.1 \\
$H_{7k}$&27.3 &35.7 &43.3 &46.9 &50.2 &54.4 &\textbf{56.1}\dag &58.9 \\
$H_{8k}$&29.2 &38.8 &44.5 &47.9 &51.8 &54.6 &56.8 &\textbf{59.5}\dag \\
\hline
\hline
Mixed&-&23.4*&32.5*&39.3*&45.5*&51.0*&55.1*&59.3*\\
\hline

\end{tabular}
\end{small}
\end{center}
\vspace{-4mm}
\caption{\textbf{Staged-Training(H-L) vs. Mixed-training.} Here ($H_{Xk}$, $L_{Yk}$) means the first stage of Staged-Training uses $Xk$ high-res images and the second stage uses $Yk$ low-res images. Mixed Training uses combined data in equal proportion. \dag and * indicate numbers in each column that can be compared directly. 
}
\label{tab:seqmix}
\vspace{-0.2in}
\end{table}

\vspace{-0.1in}
\subsection{Results on Caltech-UCSD Birds-200-2011 dataset}

\begin{table}[t]
\begin{center}
\begin{small}

\begin{tabular}{c|c|c|c}

\multirow{2}{*}{Train/Test Strategy} & \multirow{2}{*}{Train} &  Test on &Test on\\

&&High-res&Low-res\\
\hline
``AlexNet''& High& 67.6 &21.1\\
``AlexNet'' & Low & 36.7 & 51.3\\
Mixed-Training & High+Low & 61.2 &53.6\\
Super-Res NBSRF\cite{Salvador2015} & Low &-&50.1 \\
Staged-Training(L-H) & 1.Low 2.High & 56.8 & 36.3 \\
Staged-Training(H-L) & 1. High 2.Low &58.1 & \textbf{55.3}

\end{tabular}

\end{small}
\end{center}
\vspace{-4mm}
\caption{\textbf{Results on Birds-200-2011} 
Accuracy for traditional training (``AlexNet''), several baseline methods, and our Staged-Training method. While we target the low-res test scenario, we also show results for high-res test for comparison. 
}
\label{tab:Caltech-CUB}
\vspace{-0.1in}
\end{table}

We further test our method and baselines on the Caltech-UCSD Birds dataset(CUB-200-2011)~\cite{wah2011caltech}. 
The bird images are natural images taken in the wild, and are on average lower in resolution than the cars data, so they contain less fine-detail information. 

The results in Table \ref{tab:Caltech-CUB} reveal that our method, staged training, again outperforms ``AlexNet'', mixed training and the super-resolution baseline on low-res test data. Staged training  boosts the accuracy of fine-grained classification for birds from 51.3\% to 55.3\%, while mixed training obtains a lower accuracy of 53.6\%. The ``oracle'' performance of high-res training and testing is 67.6\%. This is lower than state-of-the-art on this dataset because our base network is much simpler, however, we expect our results to generalize to deeper networks. Our implementation of the LR-CNN proposed in \cite{chevalierlr} gets an accuracy of 23.6\%. 
These results demonstrate that transferring knowledge from high-res to low-res data can improve performance on a variety of fine-grained category problems.

\section{Conclusion}
\label{conclusion}

In this paper, we proposed a simple but effective staged training scheme to learn powerful CNN filters for fine-grained classification of low-res test data. Our extensive experiments on Stanford Car dataset \cite{stanfordcar} and Caltech-UCSD Birds dataset \cite{wah2011caltech} demonstrate that staged training outperforms multiple baselines, including the simple ``AlexNet'', BB-3D-G \cite{stanfordcar} and NBSRF \cite{Salvador2015}. We believe our method is general and can be applied to other network structures.

\section{Acknowledgement}
This work was supported in part by NSF award IIS-1535797 and IIS-1451244.

\bibliographystyle{IEEEbib}
\bibliography{strings,refs,egbib}

\end{document}